\documentclass[10pt, a4paper]{article}
\usepackage[utf8]{inputenc}
\usepackage[margin=1in]{geometry}
\usepackage{amsmath, amssymb, amsfonts}
\usepackage{graphicx}
\usepackage[hypertexnames=false]{hyperref}
\usepackage{titlesec}
\usepackage{booktabs}
\usepackage{fancyhdr}
\usepackage{float}
\usepackage[ruled,vlined]{algorithm2e}
\graphicspath{{assets/}}

\titlespacing*{\section}{0pt}{10pt}{5pt}
\titlespacing*{\subsection}{0pt}{6pt}{3pt}
\title{\textbf{Isokinetic Flow Matching for Pathwise Straightening of Generative Flows}}
\author{Tauhid Khan}
\date{}

\begin{document}
\maketitle
\vspace{-1.2cm}




\section*{Abstract}
Flow Matching (FM) constructs linear conditional probability paths, but the learned \emph{marginal} velocity field inevitably exhibits strong curvature due to trajectory superposition. This curvature severely inflates numerical truncation errors, bottlenecking few-step sampling. To overcome this, we introduce \textbf{Isokinetic Flow Matching (Iso-FM)}, a lightweight, \emph{Jacobian-free} dynamical regularizer that directly penalizes pathwise acceleration. By using a self-guided finite-difference approximation of the material derivative $Dv/Dt$, Iso-FM enforces local velocity consistency without requiring auxiliary encoders or expensive second-order autodifferentiation. Operating as a pure \emph{plug-and-play} addition to single-stage FM training, Iso-FM dramatically improves few-step generation. On CIFAR-10 (DiT-S/2), Iso-FM slashes conditional non-OT FID@2 from 78.82 to 27.13—a \textbf{2.9$\times$ relative efficiency gain}—and reaches a best-observed FID@4 of 10.23. These results firmly establish acceleration regularization as a principled, compute-efficient mechanism for fast generative sampling.

\begin{figure}[t]
\centering
\begin{minipage}[t]{0.49\linewidth}
\centering
\includegraphics[width=\linewidth]{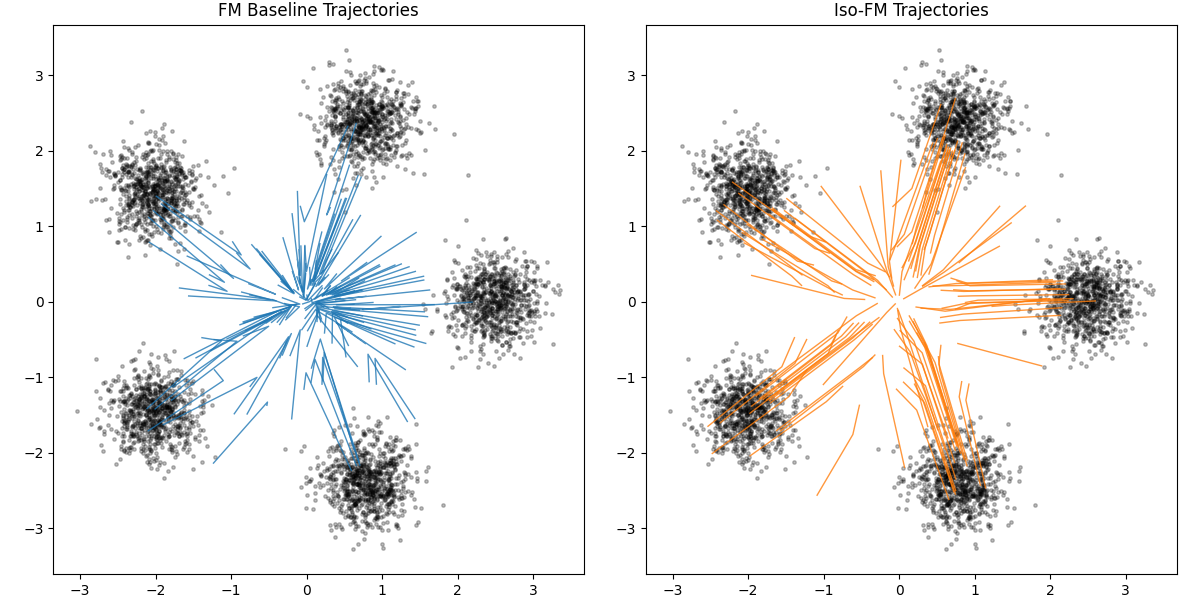}
\vspace{2pt}
{\footnotesize (a) Baseline FM vs Iso-FM trajectories}
\end{minipage}
\hfill
\begin{minipage}[t]{0.49\linewidth}
\centering
\includegraphics[width=\linewidth]{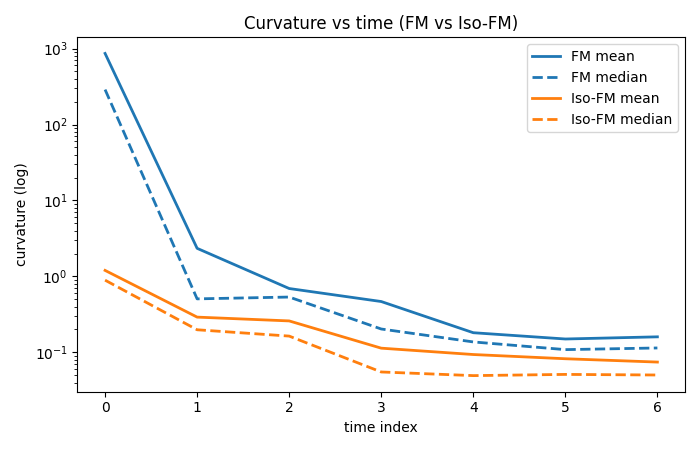}
\vspace{2pt}
{\footnotesize (b) Curvature comparison}
\end{minipage}
\caption{Teaser: Iso-FM suppresses path bending in low-dimensional transport. Left: baseline FM trajectories are more curved than Iso-FM trajectories. Right: Iso-FM yields lower curvature, consistent with reducing the material acceleration $Dv/Dt$.}
\label{fig:teaser_straightening}
\end{figure}

\section{Introduction}
This paper studies sampling efficiency in flow-based generative modeling through transport dynamics rather than model scaling. Our premise is that few-step quality is governed by the \emph{dynamical straightness} of the learned velocity field. The key quantity is the material derivative
\begin{equation}
\label{eq:intro_material_derivative}
\frac{Dv}{Dt}=\partial_t v + (v\cdot\nabla_x)v,
\end{equation}
which measures acceleration along trajectories. When $\|Dv/Dt\|$ is large, coarse ODE integration incurs large truncation error; when it is small, trajectories are closer to constant-velocity transport and are easier to integrate with few NFEs.

Iso-FM targets this mechanism with a Jacobian-free regularizer implemented by one lookahead model evaluation and stop-gradient matching, while preserving the standard FM objective and single-stage training loop. We evaluate Iso-FM as a controlled augmentation to FM (not as an absolute SOTA claim): under matched DiT-S/2 CIFAR-10 settings, conditional non-OT FID@2 improves from 78.82 to 27.13 (2.9$\times$ relative few-step efficiency gain).
Figure~\ref{fig:teaser_straightening} previews this mechanism: baseline FM trajectories bend more and exhibit higher curvature, while Iso-FM yields straighter, lower-curvature transport.

\textbf{Our main contributions are summarized as follows:}
\begin{itemize}
    \item \textbf{Dynamical Regularization for Flow Matching:} We establish a direct theoretical link between the material derivative ($Dv/Dt$) of the learned Eulerian velocity field and the truncation error experienced during few-step ODE integration.
    \item \textbf{Jacobian-Free Optimization:} We introduce Isokinetic Flow Matching (Iso-FM), a lightweight, plug-and-play regularizer that penalizes pathwise acceleration. We propose a self-guided finite-difference approximation that requires only standard forward evaluations and stop-gradient targets, entirely avoiding expensive Jacobian-vector products.
    \item \textbf{Improved Few-Step Efficiency:} We empirically demonstrate that Iso-FM shifts the Pareto front for single-stage Flow Matching. On CIFAR-10 with a DiT-S/2 backbone, Iso-FM achieves a 2.9$\times$ relative efficiency gain at NFE=2 (FID 78.82 $\rightarrow$ 27.13), proving that acceleration regularization effectively straightens generative flows without requiring multi-stage distillation.
\end{itemize}

\section{Related Work}

\subsection{Flow Matching and Eulerian Generative Models}

Generative flow models aim to construct a time-dependent vector field $v_t(x)$ that pushes a simple prior distribution $p_0$ (e.g., a standard Gaussian) to a complex data distribution $p_1$. The evolution of the probability density $p_t(x)$ along this vector field is governed by the continuity equation:
\begin{equation}
\label{eq:continuity_equation}
\partial_t p_t + \nabla \cdot (p_t v_t) = 0.
\end{equation}
The ideal training objective would directly match the neural network $v_\theta(x, t)$ to the true marginal vector field $u_t(x)$ that generates the target probability path $p_t(x)$:
\[
\mathcal{L}_{marginal} = \mathbb{E}_{t \sim \mathcal{U}(0,1), x \sim p_t(x)} [\| v_\theta(x, t) - u_t(x) \|^2].
\]
However, computing the marginal vector field $u_t(x)$ and sampling directly from the marginal path $p_t(x)$ are computationally intractable for complex datasets. 

Conditional Flow Matching (CFM) \cite{lipman2023flowmatchinggenerativemodeling} bypasses this intractability by conditioning the vector field on individual source-target pairs. Given a source sample $x_0 \sim p_0$ and a target data sample $x_1 \sim p_1$, we can construct a simple conditional probability path $p_t(x | x_0, x_1)$. Under the Optimal Transport (OT) formulation, this path is defined as a linear interpolation:
\begin{equation}
\label{eq:ot_path}
x_t = (1-t)x_0 + t x_1.
\end{equation}
By differentiating Equation~\ref{eq:ot_path} with respect to time, the corresponding conditional vector field is simply the constant velocity between the two endpoints:
\begin{equation}
\label{eq:conditional_velocity}
u_t(x_t | x_0, x_1) = x_1 - x_0.
\end{equation}
The fundamental theorem of Flow Matching demonstrates that regressing against this easily computable conditional vector field provides identical gradients to the intractable marginal objective. This yields the standard FM loss:
\begin{equation}
\label{eq:fm_objective_related}
\mathcal{L}_{FM} = \mathbb{E}_{t, x_0, x_1, x_t} [\| v_\theta(x_t, t) - (x_1 - x_0) \|^2].
\end{equation}
While the individual conditional trajectories (Equation~\ref{eq:ot_path}) are perfectly straight, the neural network $v_\theta$ does not have access to the endpoints during inference. Instead, it learns to approximate the \emph{marginal} velocity field, which is the expectation over all possible conditional fields that pass through a given point $x$ at time $t$:
\begin{equation}
\label{eq:marginal_velocity}
v(x,t) = \mathbb{E}_{x_0, x_1 \sim p(x_0, x_1 | x_t)} [x_1 - x_0].
\end{equation}
Because independent couplings of $x_0$ and $x_1$ cause many distinct conditional trajectories to intersect in the state space, this expectation averages out conflicting velocity directions. As a direct consequence, the resulting marginal vector field $v(x,t)$ exhibits strong geometric curvature, necessitating multi-step numerical integration during sampling. This is a fundamental bottleneck for few-step generation, and motivates the need for dynamical regularization to suppress pathwise acceleration.

\subsection{Flow Straightening and Acceleration Control}
Several works address trajectory curvature to enable faster sampling. \emph{Rectified Flow} iteratively refines the coupling $(Z_0, Z_1)$ by using samples from a pre-trained model as new data pairs, effectively "straightening" the flow via recursive training \cite{liu2022flowstraightfastlearning}.
Similarly, Optimal Transport Conditional Flow Matching (OT-CFM) minimizes trajectory curvature by matching source and target samples within a minibatch using optimal transport, creating simpler and straighter ground-truth paths for the model to learn \cite{tong2024improving}.
In \emph{Minimizing Trajectory Curvature of ODE-based Generative Models}, Lee et al.\ explicitly connect solver error to path curvature and propose minimizing trajectory curvature by untangling path intersections through improved data-noise pairing with a $\beta$-VAE-style encoder \cite{lee2023minimizingtrajectorycurvatureodebased}.
More recently, \emph{OAT-FM} applies optimal transport constraints to minimize acceleration $\mathbb{E}[\|\ddot{x}\|^2]$ as a post-training refinement step \cite{yue2025oatfmoptimalaccelerationtransport}.
\emph{Shortcut Models} generalize this by conditioning the network on a step-size parameter $d$, learning a "shortcut" $s_\theta(x_t, t, d)$ that predicts the state $x_{t+d}$ directly. They train via a self-consistency recurrence where a step of size $2d$ must match two steps of size $d$: $s(x_t, t, 2d) \approx \frac{1}{2}(s(x_t, t, d) + s(x_{t+d}, t+d, d))$ \cite{frans2025stepdiffusionshortcutmodels}. Unlike Iso-FM, Shortcut Models require variable-step conditioning and do not strictly enforce local stationarity of the velocity field itself.
While these methods successfully straighten paths by modifying source-target coupling (a Lagrangian approach), Iso-FM achieves straightness in the Eulerian regime by regularizing the learned velocity field directly, without auxiliary encoders or requiring near-optimal couplings.

\subsection{Flow Map and Consistency Methods}
Consistency Models (CM) \cite{song2023consistencymodels} and Flow Map Matching (FMM) \cite{boffi2025flowmapmatchingstochastic} learn the \emph{Lagrangian solution operator} $\Phi_{s \to t}(x)$ directly. The core objective enforces that points along a trajectory map to the same origin or endpoint:
\begin{equation}
\label{eq:cm_objective_related}
\mathcal{L}_{CM} = \mathbb{E}[\| f_\theta(x_{t+\Delta t}, t+\Delta t) - f_\theta(x_t, t) \|^2].
\end{equation}
Recent extensions build upon these concepts to straighten flows.
\emph{Consistency Flow Matching} (CFM) \cite{yang2024consistencyflowmatchingdefining}
enforces self-consistency in the velocity field using a hybrid objective:
a velocity-consistency penalty plus a terminal-consistency loss that encourages
shared endpoint prediction (e.g., $x_1$) along trajectories.
It also evaluates consistency between $t$ and $t+\Delta t$ with an EMA target
network, inheriting the optimization dynamics of consistency training.
Similarly, \emph{MeanFlow} \cite{geng2025meanflowsonestepgenerative}
modifies the regression target by introducing an average velocity
\begin{equation*}
U_{s \to t} = \frac{1}{t-s}\int_s^t v_\tau d\tau,
\end{equation*}
and enforcing an identity that relates $v_t$ to $\frac{d}{dt}U$.

\textbf{Position of Iso-FM.}
Iso-FM distinguishes itself by operating purely in the \emph{Eulerian} regime without learning an explicit integral map $\Phi$ or conditioning on step size $d$. Instead, it regularizes the \emph{material derivative} $Dv/Dt$ of the velocity field directly using a finite-difference approximation.
Compared with Consistency Flow Matching \cite{yang2024consistencyflowmatchingdefining}, which emphasizes cross-time velocity consistency as a primary training paradigm, Iso-FM is used as a \emph{plug-and-play, Jacobian-free regularizer} on top of standard FM regression. In this sense, Iso-FM is orthogonal rather than competitive in setup: it preserves single-stage FM training and can be composed with broader flow-straightening pipelines.
Iso-FM occupies a distinct position: it neither replaces the FM objective nor learns solution maps. Instead, it modifies the \emph{geometry of the learned velocity field} by suppressing pathwise acceleration, thereby making the induced ODE integrator-friendly.

\section{Background: Eulerian and Lagrangian Perspectives}

Continuous-time generative models can be analyzed through two complementary mathematical lenses, borrowed from fluid dynamics: the Eulerian and the Lagrangian perspectives \cite{boffi2025flowmapmatchingstochastic, lipman2023flowmatchinggenerativemodeling}. Understanding this duality is crucial for contextualizing the computational bottlenecks of current models and justifying the design of Isokinetic Flow Matching.

\textbf{The Eulerian Perspective (Instantaneous Velocity).}
The Eulerian viewpoint observes the flow at fixed spatial coordinates over time. In generative modeling, this corresponds to learning a time-dependent vector field $v(x,t)$ that dictates the instantaneous rate of change of a sample. Data generation is performed by solving the corresponding Initial Value Problem (IVP) through numerical integration:
\begin{equation}
\label{eq:eulerian_ode_background}
\frac{dx(t)}{dt} = v(x(t),t), \quad x(0) = x_0.
\end{equation}
Flow Matching \cite{lipman2023flowmatchinggenerativemodeling} is fundamentally Eulerian. It trains a neural network $v_\theta(x,t)$ to regress the instantaneous target velocity at specific coordinates. While Eulerian objectives are stable and simple to optimize in a single stage, they suffer during inference: if the learned vector field contains complex geometry or crossing paths, the ODE solver requires many small steps (high NFE) to bound the discretization error \cite{lee2023minimizingtrajectorycurvatureodebased}.

\textbf{The Lagrangian Perspective (Particle Trajectory).}
Conversely, the Lagrangian viewpoint tracks the continuous trajectory of individual particles (samples) as they move through the state space. Mathematically, this is formalized by the solution operator or flow map $\Phi$, which maps an initial state directly to a future state:
\begin{equation}
\label{eq:flow_map_background}
\Phi_{s \to t}(x_s) = x_t.
\end{equation}
Recent acceleration techniques, such as Consistency Models \cite{song2023consistencymodels} and Flow Map Matching \cite{boffi2025flowmapmatchingstochastic}, adopt a Lagrangian approach. By training a network to explicitly approximate $\Phi_{s \to t}$, they allow for one-step or few-step sampling by directly jumping across time. However, learning global Lagrangian flow maps is notoriously unstable and difficult, often requiring complex multi-stage distillation pipelines, exponential moving average (EMA) teacher networks, and carefully tuned discretization schedules \cite{song2023consistencymodels, yang2024consistencyflowmatchingdefining}.

\textbf{Bridging the Gap: The Role of Pathwise Acceleration.}
We introduce this dichotomy to highlight the core theoretical motivation of Iso-FM. The transition between the Eulerian vector field and the Lagrangian trajectory is governed by \emph{acceleration}. If an Eulerian velocity field has zero acceleration along its own flow lines, a particle's velocity remains constant, and the corresponding Lagrangian flow map becomes trivially linear:
\begin{equation}
\label{eq:linear_flow_map}
\Phi_{0 \to t}(x_0) = x_0 + t \cdot v(x_0, 0).
\end{equation}
This insight dictates our methodology. Rather than abandoning the stable Eulerian training paradigm to learn explicit Lagrangian flow maps, Iso-FM operates purely in the Eulerian regime. By directly suppressing the pathwise acceleration of the vector field, Iso-FM implicitly induces near-linear Lagrangian transport maps. This approach allows us to approach the few-step sampling efficiency characteristic of Lagrangian methods while retaining the lightweight, single-stage training dynamics of Eulerian Flow Matching.

\textbf{Key observation:}
If the Eulerian velocity field has \emph{zero (or small) acceleration along trajectories}, then the flow map becomes approximately linear:
\begin{equation}
\label{eq:linearized_flow_map_background}
\Phi_t(x_0) \approx x_0 + t\, v(x_0,0).
\end{equation}

Iso-FM exploits this fact: by straightening the Eulerian field, it implicitly induces simple Lagrangian transport maps without learning them explicitly.

\section{Isokinetic Flow Matching (Iso-FM)}

\subsection{Dynamical Motivation: Acceleration and Sampling Error}

Let $x(t)$ follow the learned ODE:
\begin{equation}
\label{eq:dyn_ode}
\dot{x}(t) = v(x(t),t).
\end{equation}

The acceleration of a particle moving through the field is given by the \textbf{material derivative}:
\begin{equation}
\label{eq:dyn_material_derivative}
\frac{Dv}{Dt}
=
\frac{\partial v}{\partial t}
+
(v \cdot \nabla_x)v.
\end{equation}

A Taylor expansion of the trajectory yields:
\begin{equation}
\label{eq:dyn_taylor}
x(1) = x(0) + v(x_0,0) + \frac{1}{2}\frac{Dv}{Dt}(x_0,0) + \mathcal{O}(\cdot).
\end{equation}

Thus, one-step Euler error is dominated by the acceleration term. High curvature and acceleration force numerical solvers to take many small steps.

\textbf{Conclusion:}
Few-step generation requires suppressing $\frac{Dv}{Dt}$ along trajectories.

\subsection{The MeanFlow Identity (Optional Connection)}
MeanFlow-style formulations introduce interval-averaged velocity $U_{s\to t} = \frac{1}{t-s}\int_s^t v_\tau d\tau$ and relate it to instantaneous velocity through
\begin{equation}
\label{eq:meanflow_identity}
v_t = U_{s\to t} + (t-s)\frac{d}{dt}U_{s\to t}.
\end{equation}
This identity motivates temporal consistency objectives. In contrast, Iso-FM directly regularizes the local material acceleration of the Eulerian field via a finite-difference surrogate, while preserving the original FM regression target.

\subsection{Practical Self-Guided Iso-FM}

Iso-FM replaces explicit derivatives with a \textbf{self-guided finite-difference approximation}. For a small $\varepsilon > 0$, define a forward Euler step using the model’s own velocity:
\begin{equation}
\label{eq:isofm_lookahead}
x_{t+\varepsilon} = x_t + \varepsilon\, v_\theta(x_t,t).
\end{equation}

Acceleration is approximated by enforcing temporal consistency of velocity predictions:
\begin{equation}
\label{eq:isofm_loss_simple}
\mathcal{L}_{\text{Iso}}
=
\left\|
v_\theta(x_t,t)
-
\text{sg}\!\left[
v_\theta(x_{t+\varepsilon}, t+\varepsilon)
\right]
\right\|^2.
\end{equation}

This loss penalizes changes in velocity \emph{along the model’s own trajectory}, directly suppressing convective acceleration.

\paragraph{Algorithmic Training Step.}
Algorithm~\ref{alg:isofm_formal} summarizes the training step and is placed here
to directly accompany the practical self-guided objective.
The implementation follows \texttt{isofm/core/flow.py}.

\begin{algorithm}[H]
\SetAlgoLined
\DontPrintSemicolon
\SetKwInput{Input}{Input}
\Input{Data $x_1$, noise $x_0 \sim \mathcal{N}(0, I)$, model $v_\theta$, weights $\lambda_{FM}, \lambda_{Iso}$, exponent $\alpha$}
\BlankLine
$t \sim \text{LogitNormal}(0, 1)$ \tcp*{Sample time step}
$x_t = (1-t)x_0 + t x_1$ \tcp*{Conditional path}
$u_t = x_1 - x_0$ \tcp*{Target velocity}
\BlankLine
\tcp{Standard Flow Matching Branch}
$v_{curr} = v_\theta(x_t, t)$ \;
$\mathcal{L}_{FM} = \| v_{curr} - u_t \|^2_2$ \;
\BlankLine
\tcp{Isokinetic Regularization (Jacobian-Free)}
$\varepsilon \sim p(\varepsilon)$ \tcp*{Sample small lookahead step}
$x_{t+\varepsilon} = x_t + \varepsilon \cdot \text{stop\_gradient}(v_{curr})$ \;
$v_{next} = v_\theta(x_{t+\varepsilon}, t+\varepsilon)$ \;
\BlankLine
$w = (1-t)^\alpha / \varepsilon$ \tcp*{Temporal weighting}
$s = \|\text{stop\_gradient}(v_{curr})\|_2 + \zeta$ \tcp*{Stability constant $\zeta$}
$\mathcal{L}_{Iso} = w \cdot \left\| \frac{v_{curr} - \text{stop\_gradient}(v_{next})}{s} \right\|_1$ \;
\BlankLine
\Return $\mathcal{L} = \lambda_{FM}\mathcal{L}_{FM} + \lambda_{Iso}\mathcal{L}_{Iso}$
\caption{Isokinetic Flow Matching (Iso-FM) Training Step}
\label{alg:isofm_formal}
\end{algorithm}

\subsection{Optional OT Path Coupling}

Following the minibatch Optimal Transport Conditional Flow Matching (OT-CFM) framework \cite{tong2024improving}, for OT-enabled runs we apply minibatch optimal transport coupling before building the FM interpolation path. Given Gaussian samples $\{x_0^{(i)}\}_{i=1}^B$ and normalized data $\{x_1^{(j)}\}_{j=1}^B$, we solve a linear assignment problem:
\begin{equation}
\label{eq:ot_assignment}
\pi^*
=
\arg\min_{\pi \in S_B}
\sum_{i=1}^{B}
\left\|
x_0^{(i)} - x_1^{(\pi(i))}
\right\|_2^2.
\end{equation}
We compute the pairwise squared Euclidean cost matrix and solve the assignment with the Hungarian algorithm, then reorder data and labels using $\pi^*$. Flow Matching and Iso-FM objectives are then applied on
\begin{equation}
\label{eq:ot_interpolation}
x_t = (1-t)x_0 + t\,x_1^{(\pi^*)}.
\end{equation}
This isolates the effect of improved source-target pairing from the Iso-FM dynamical regularization term.

\subsection{Interpretation and Path Geometry}

Iso-FM does not constrain endpoints or enforce map consistency.
Instead, it:
\begin{itemize}
\item straightens the Eulerian velocity field,
\item reduces kinetic energy variation,
\item and implicitly linearizes the induced flow map.
\end{itemize}

\paragraph{Low-Dimensional Path Geometry Analysis.}
To provide mechanistic evidence for path straightening, we include a controlled low-dimensional comparison between baseline FM and Iso-FM. Let trajectories evolve under the learned velocity field
\begin{equation}
\label{eq:path_geometry_dynamics}
\dot{x}(t)=v_\theta(x(t),t),\qquad \ddot{x}(t)=\frac{Dv_\theta}{Dt}=\partial_t v_\theta + (v_\theta\cdot\nabla_x)v_\theta.
\end{equation}
Iso-FM directly penalizes temporal velocity variation along model trajectories, which reduces $\|Dv_\theta/Dt\|$ and therefore suppresses path bending.

For geometric quantification, we report a curvature proxy
\begin{equation}
\label{eq:curvature_proxy}
\kappa(t)=\frac{\|\ddot{x}(t)\|}{\|\dot{x}(t)\|^2+\varepsilon},
\end{equation}
where $\varepsilon>0$ ensures numerical stability at low speed. Since $\ddot{x}(t)$ is the material acceleration, lower acceleration under Iso-FM implies lower curvature and a straighter transport map.

\begin{figure}[H]
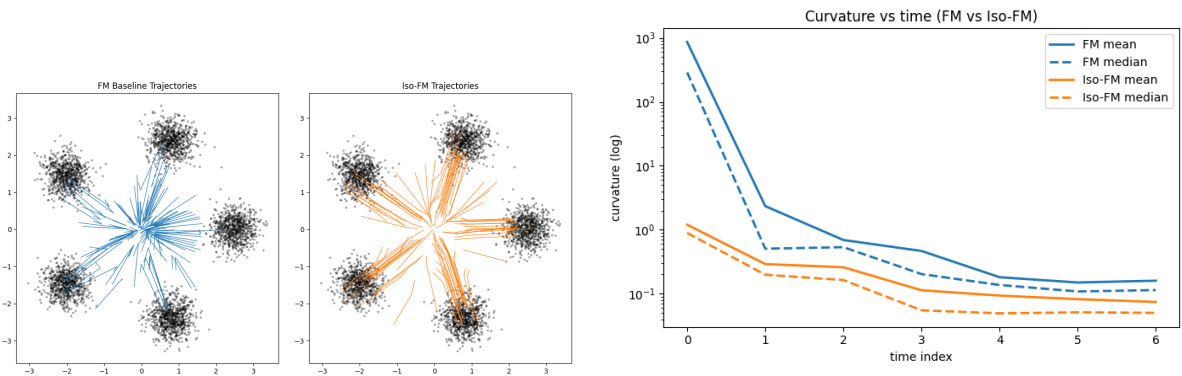

\centering
\begin{minipage}[t]{0.49\textwidth}
    \centering
    \includegraphics[width=\linewidth]{trajectories_comparison.png}
\end{minipage}
\hfill
\begin{minipage}[t]{0.49\textwidth}
    \centering
    \includegraphics[width=\linewidth]{curvature_comparison.png}
\end{minipage}
\caption{Low-dimensional geometric diagnostics of learned flows. Left (trajectory comparison): baseline FM paths exhibit stronger bending and non-uniform directional changes, while Iso-FM trajectories are more linear between source and target regions. Right (curvature comparison): the Iso-FM model maintains consistently lower curvature magnitude across integration time, empirically supporting the theoretical objective of acceleration suppression.}
\label{fig:low_dim_path_geometry}
\end{figure}

Importantly, Iso-FM incurs minimal memory overhead and integrates seamlessly into standard FM training.

\subsection*{Terminology}
The term \textit{isokinetic} denotes approximately preserved speed along trajectories. Directional changes are allowed; rapid acceleration and curvature are discouraged.

\section{Experiments}

Experiments use a DiT-S/2 backbone on CIFAR-10 across conditional/unconditional and OT/non-OT settings. In OT runs, source-target pairing is computed with minibatch Hungarian matching on squared Euclidean cost before FM interpolation. Metrics are recorded every 250 epochs. We report only the best score per metric, together with its corresponding epoch.

\textbf{Evaluation objective and claim scope.}
Our experimental objective is not to establish a new absolute state-of-the-art on CIFAR-10; rather, it is to isolate and quantify the effect of Eulerian acceleration regularization on sampling speed under controlled single-stage training. Accordingly, all comparisons are made between FM baselines and Iso-FM-augmented models with matched architecture, training horizon, and evaluation protocol.

\subsection{Training Configuration}

Table~\ref{tab:training_config} summarizes the exact optimization and model settings used for the CIFAR-10 runs.

\begin{table}[h]
\centering
\scriptsize
\caption{Training configuration for the main CIFAR-10 experiment.}
\label{tab:training_config}
\begin{tabular}{p{0.30\linewidth}p{0.63\linewidth}}
\toprule
Item & Value \\
\midrule
Backbone & DiT-S/2 Transformer (depth $=12$, hidden dim $=384$, heads $=6$, patch size $=2$, MLP ratio $=4.0$). \\
Trainable parameters & 32,485,260 (conditional, $num\_classes=10$);\\
& 32,481,804 (unconditional, $num\_classes=1$). \\
Data / resolution & CIFAR-10 train split, RGB $32\times 32$, random horizontal flip augmentation. \\
Batch size / epochs & 256 / 2500. \\
Optimizer & AdamW ($lr=5\times10^{-4}$, weight decay $=10^{-4}$). \\
Learning-rate schedule & Cosine annealing with $T_{\max}=2500$, $\eta_{\min}=0.1\times lr$. \\
EMA / precision / clipping & EMA decay $=0.9999$, bfloat16 mixed precision, gradient clipping at global norm 1.0. \\
Sampling of $t$ & Logit-normal sampling ($\mu=0.0$, $\sigma=1.0$). \\
Loss weights & $\lambda_{FM}=1.0$ for all runs; FM baseline uses $\lambda_{Iso}=0.0$, $p_{iso}=0.0$; Iso-FM uses $\lambda_{Iso}=4.0$, $p_{iso}=1.0$, $\alpha=2.0$. \\
CFG settings & Conditional: label-drop probability $0.15$ and eval CFG scales $\{2.5,3.5,4.5\}$; Unconditional: CFG scale $=1.0$. \\
OT setting & When enabled, minibatch coupling uses Hungarian assignment on squared Euclidean cost before interpolation. \\
Evaluation protocol & Every 250 epochs, 50,000 generated samples, FID at NFE $\{1,2,4\}$. \\
Checkpointing / sampling & Image sampling and checkpointing every 10 epochs, keep last 5 checkpoints. \\
\bottomrule
\end{tabular}
\end{table}

\begin{table}[h]
\centering
\scriptsize
\caption{Best FID scores and epochs for the main CIFAR-10 experiment (DiT-S/2).}
\begin{tabular}{llccc}
\toprule
Setup & Method & Best FID@1 (epoch) & Best FID@2 (epoch) & Best FID@4 (epoch) \\
\midrule
Conditional & FM & 245.3692 (1250) & 78.8200 (2250) & 27.2868 (1250) \\
Conditional & Iso-FM & \textbf{83.8460 (1500)} & \textbf{27.1267 (750)} & \textbf{15.5443 (1250)} \\
Conditional OT & Iso-FM & 104.3007 (750) & \textbf{17.7633 (750)} & \textbf{10.2254 (1000)} \\
Unconditional & FM & 327.1964 (500) & 98.2889 (500) & 47.7700 (500) \\
Unconditional & Iso-FM & \textbf{143.6726 (2000)} & \textbf{36.9151 (1000)} & \textbf{25.0528 (1750)} \\
Unconditional OT & FM & 249.0945 (500) & 79.6954 (500) & 42.4719 (750) \\
Unconditional OT & Iso-FM & \textbf{170.3066 (1250)} & \textbf{36.6866 (500)} & \textbf{24.0782 (1250)} \\
\bottomrule
\end{tabular}
\end{table}

\begin{figure}[H]
\centering
\begin{minipage}{0.49\linewidth}
\centering
\includegraphics[width=\linewidth]{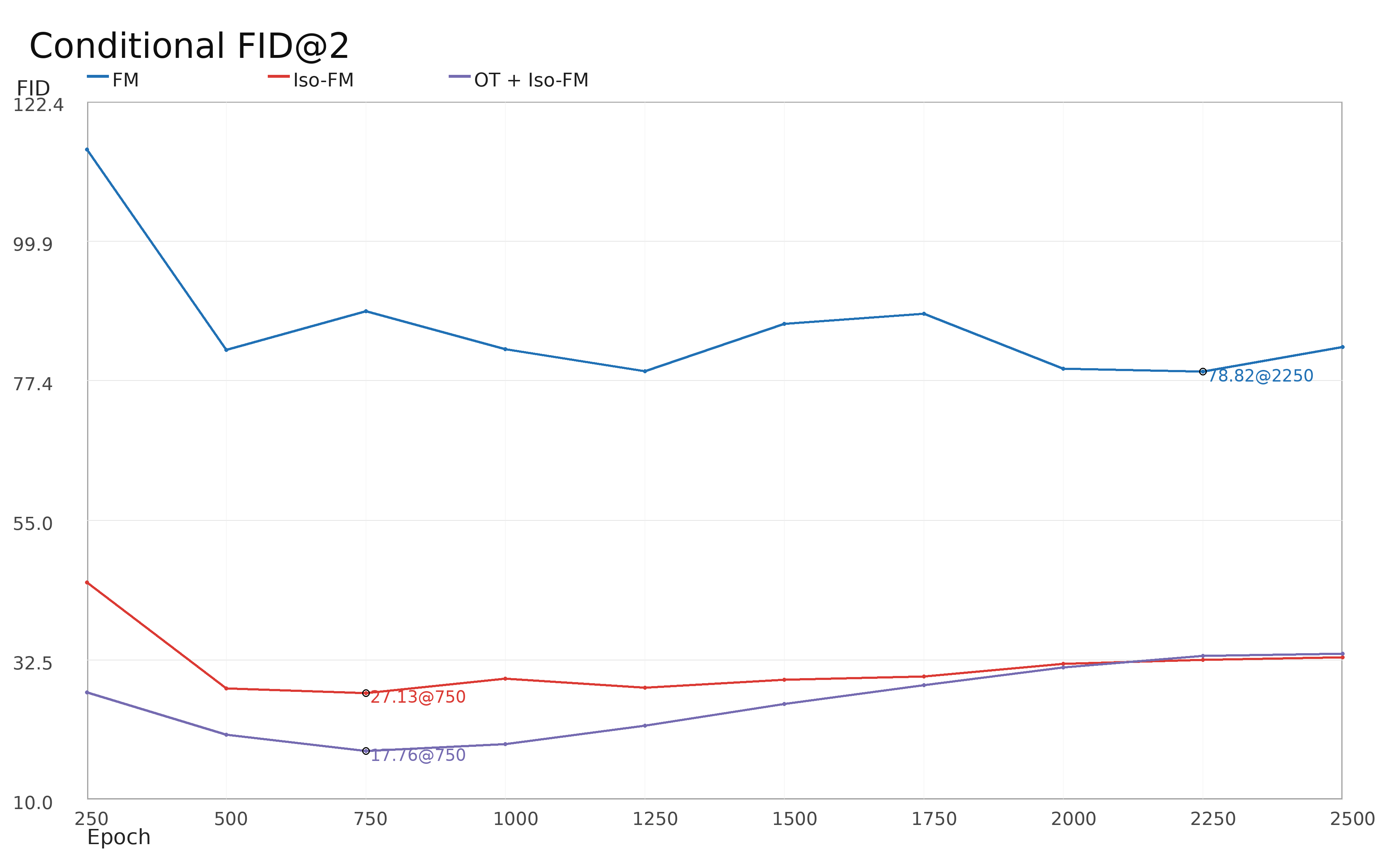}
\vspace{2pt}
{\footnotesize (a) Conditional FID@2}
\end{minipage}
\hfill
\begin{minipage}{0.49\linewidth}
\centering
\includegraphics[width=\linewidth]{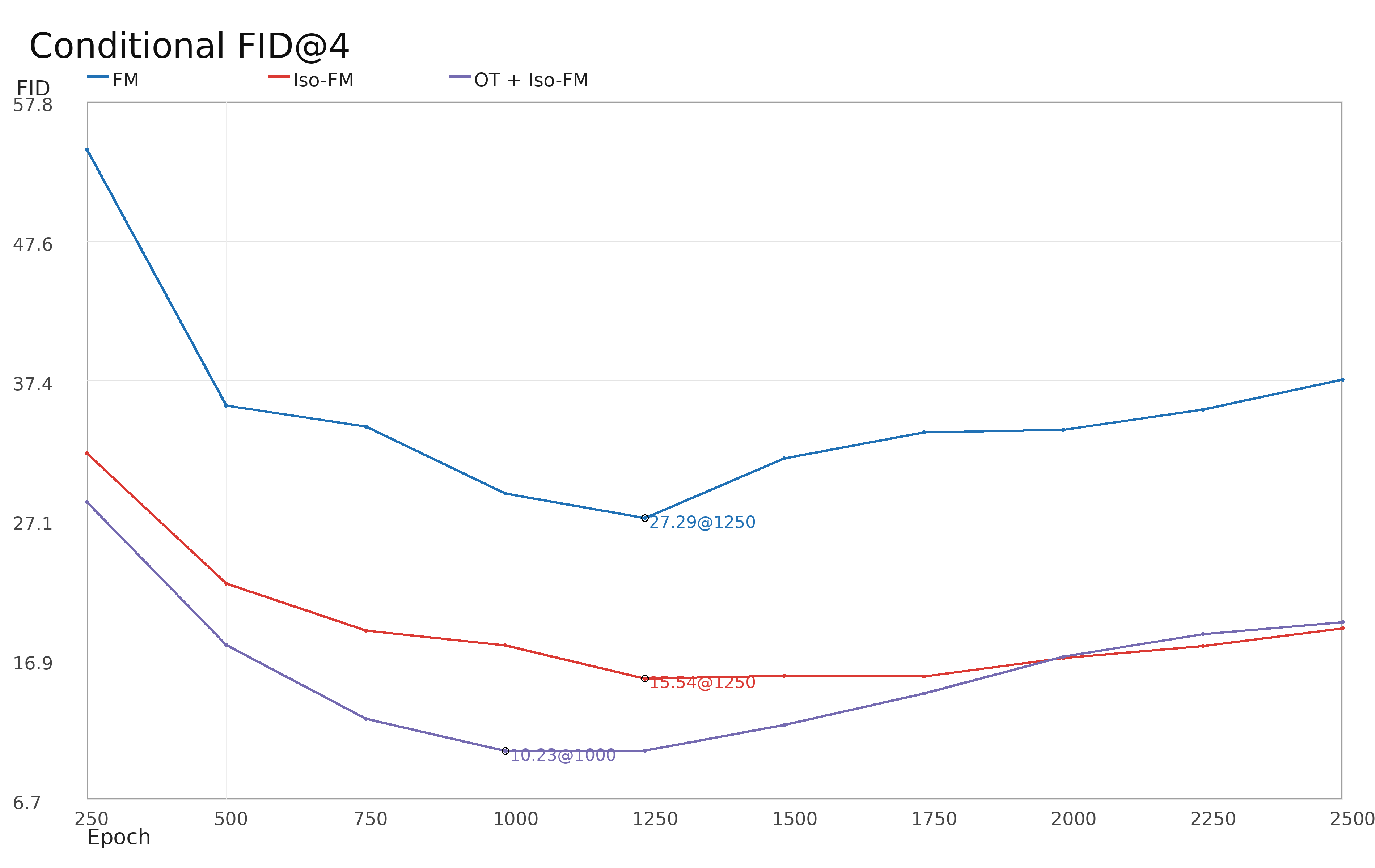}
\vspace{2pt}
{\footnotesize (b) Conditional FID@4}
\end{minipage}
\caption{Conditional CIFAR-10 training dynamics (DiT-S/2): FID@2 and FID@4 versus epoch for FM, Iso-FM, and OT+Iso-FM.}
\label{fig:conditional_curves_side_by_side}
\end{figure}

\begin{figure}[H]
\centering
\begin{minipage}{0.49\linewidth}
\centering
\includegraphics[width=\linewidth]{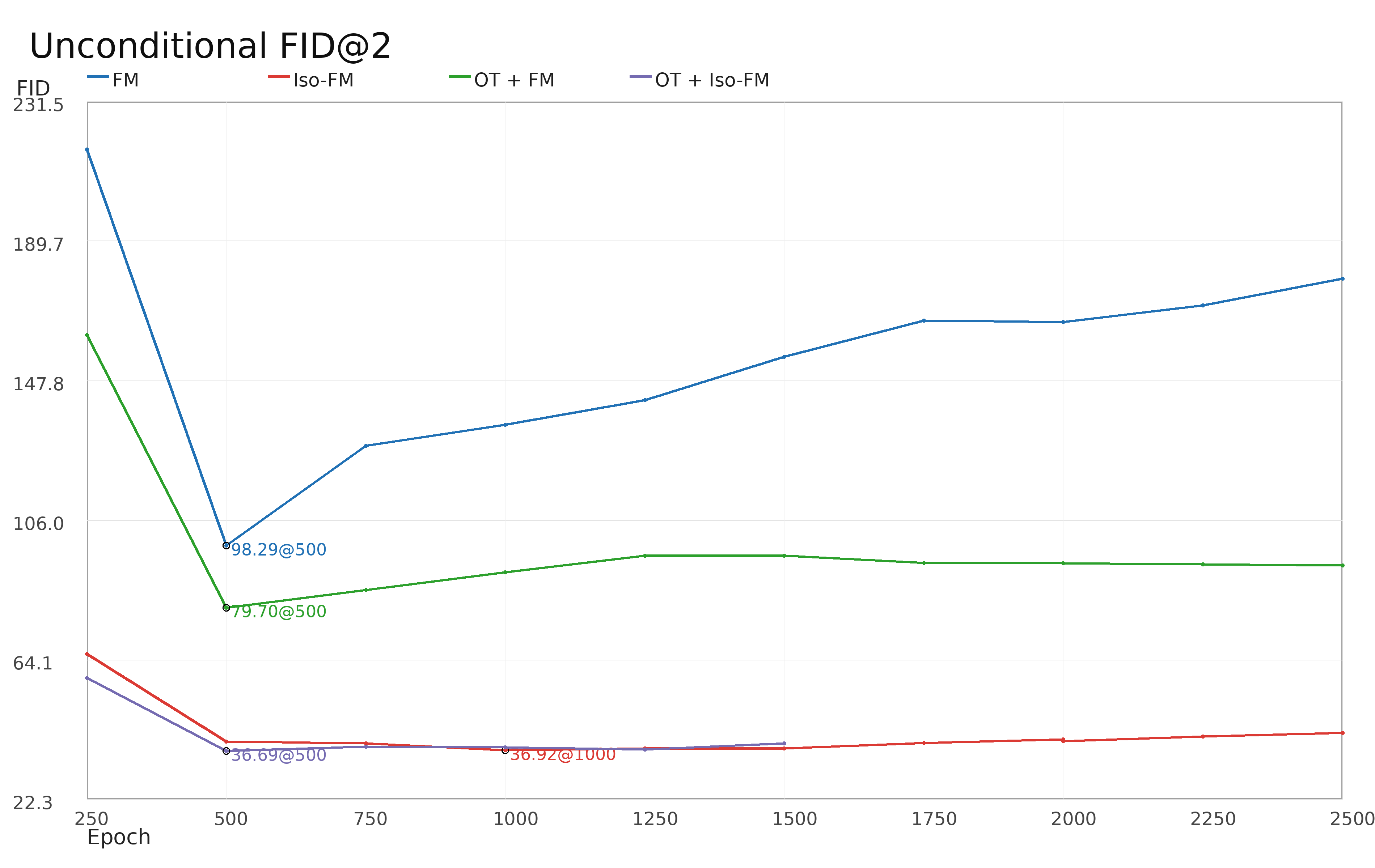}
\vspace{2pt}
{\footnotesize (a) Unconditional FID@2}
\end{minipage}
\hfill
\begin{minipage}{0.49\linewidth}
\centering
\includegraphics[width=\linewidth]{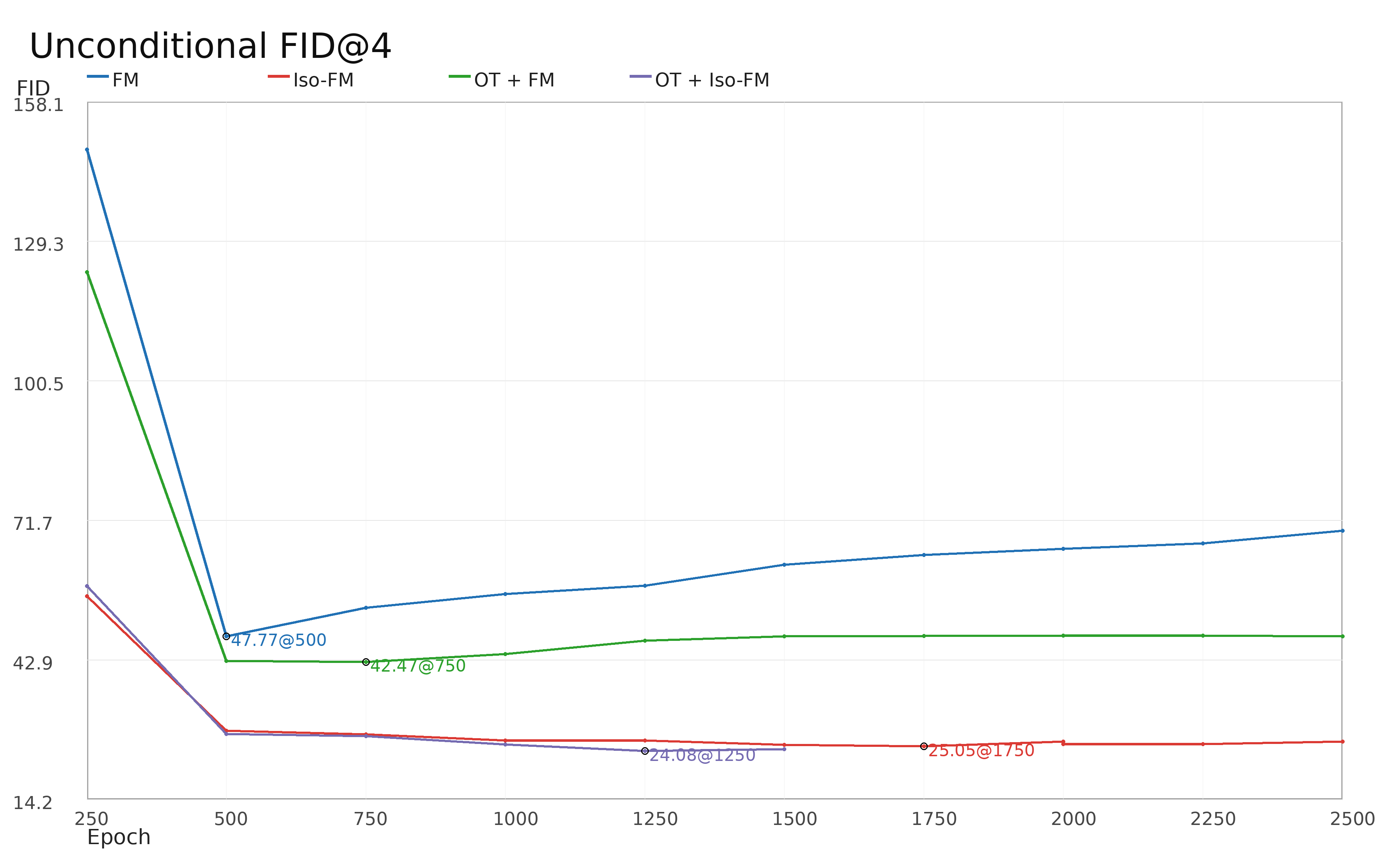}
\vspace{2pt}
{\footnotesize (b) Unconditional FID@4}
\end{minipage}
\caption{Unconditional CIFAR-10 training dynamics (DiT-S/2): FID@2 and FID@4 versus epoch for FM, Iso-FM, OT+FM, and OT+Iso-FM.}
\label{fig:unconditional_curves_side_by_side}
\end{figure}

\subsection{Analysis of Results}
Iso-FM consistently improves best low-step FID against FM in both conditional and unconditional settings. The headline conditional non-OT result at NFE$=2$ is 78.82 $\rightarrow$ 27.13, which corresponds to a \textbf{2.9$\times$ improvement in few-step efficiency} (65.58\% FID reduction). In non-OT runs, Iso-FM also improves unconditional FID@2 by 62.44\% (98.29 $\rightarrow$ 36.92). In OT runs, Iso-FM reduces best FID@2 by 53.97\% (79.70 $\rightarrow$ 36.69) and best FID@4 by 43.31\% (42.47 $\rightarrow$ 24.08) against the OT-FM baseline. These results indicate a favorable quality-vs-NFE trade-off shift for single-stage FM, even without claiming absolute leaderboard SOTA.

\section{Conclusion}

Iso-FM is a methodological contribution for improving few-step FM sampling via acceleration control. The method regularizes the material derivative in a Jacobian-free manner and is integrated as a plug-and-play term in standard single-stage FM training.

Under matched DiT-S/2 CIFAR-10 settings, Iso-FM consistently improves low-NFE performance relative to FM baselines, including a 2.9$\times$ relative efficiency gain at conditional non-OT FID@2 (78.82 $\rightarrow$ 27.13). These results indicate that pathwise acceleration regularization can shift the quality--compute trade-off favorably without requiring multi-stage distillation or architectural changes.

\subsection{Limitations and Broader Impact}
Iso-FM is a methodological contribution focused on dynamics-aware efficiency rather than an absolute SOTA benchmark claim. While we observe strong relative gains at low NFE under controlled single-stage FM training, we do not claim superiority over heavily scaled or multi-stage distilled pipelines. 

Furthermore, we acknowledge a fundamental theoretical limit to flow straightening: perfectly linear global transport maps ($\frac{Dv}{Dt} \equiv 0$) are mathematically unachievable for complex data. As derived in Appendix~\ref{app:fundamental_limit}, the material acceleration of the marginal field is strictly bounded by the divergence of the conditional velocity variance. At $t \approx 0$, the model must resolve the ambiguity of the prior by accelerating toward specific data modes. Thus, Iso-FM mitigates \emph{unnecessary} curvature but cannot eliminate the mandatory initial acceleration required for diverse, multimodal generation.

Broader validation on larger datasets, higher-resolution regimes, and stronger backbones remains necessary. Nevertheless, the Jacobian-free and plug-and-play nature of Iso-FM makes it a practical tool for reducing sampling compute and potentially lowering energy use in training-to-inference workflows. As with other generative-model acceleration methods, improved sampling efficiency can also lower misuse barriers for synthetic media generation; therefore, deployment should pair efficiency improvements with standard safeguards, provenance tooling, and policy controls.

\textbf{The Trade-off Between Straightness and Generative Variability.}
A natural question arises: does penalizing the material acceleration restrict the diversity or variability of the generated samples? As established by the fundamental limit in Appendix~\ref{app:fundamental_limit}, the marginal acceleration $\frac{Dv}{Dt}$ is inextricably linked to the conditional velocity variance $\Sigma(x,t)$, which represents the model's multimodality. A strict, hard constraint of $\frac{Dv}{Dt} \equiv 0$ everywhere would mathematically require $\Sigma$ to vanish, effectively forcing a deterministic, one-to-one mapping that would cause severe mode collapse.

However, because Iso-FM is applied as a \emph{soft} auxiliary regularizer ($\lambda_{Iso} \mathcal{L}_{Iso}$), it avoids this catastrophic loss of variability. The primary Flow Matching objective ensures the preservation of $\Sigma(x,t)$ to accurately capture the diverse data distribution. Simultaneously, the Iso-FM regularizer acts as a dynamical dampener. It allows the necessary macroscopic acceleration required for a noise sample to ``commit" to a specific data mode, while heavily penalizing the unnecessary, high-frequency convective accelerations (i.e., twisted or crossing trajectories) that inflate ODE truncation errors. Empirically, our high visual fidelity and robust FID scores confirm that Iso-FM successfully straightens the transport geometry without sacrificing the underlying diversity of the sampling process.

\clearpage
\bibliographystyle{plain}
\bibliography{references}

\clearpage
\appendix
\section*{Appendix A: Theory of Isokinetic Flow Matching}

\subsection*{A.1 Eulerian Dynamics and Material Acceleration}

Let $v(x,t): \mathbb{R}^d \times [0,1] \to \mathbb{R}^d$ be a smooth time-dependent vector field.
A particle trajectory $x(t)$ follows the ODE
\[
\frac{dx(t)}{dt} = v(x(t), t), \quad x(0) = x_0.
\]

The acceleration of the particle is the total derivative of the velocity along the trajectory:
\[
\frac{d}{dt} v(x(t), t).
\]

Applying the chain rule yields the \emph{material derivative}:
\begin{equation}
\label{eq:material_derivative}
\frac{Dv}{Dt}(x,t)
=
\frac{\partial v}{\partial t}(x,t)
+
\left( v(x,t) \cdot \nabla_x \right) v(x,t).
\end{equation}

This quantity decomposes acceleration into:
\begin{itemize}
\item \textbf{Local (temporal) acceleration} $\partial_t v$,
\item \textbf{Convective acceleration} $(v \cdot \nabla_x) v$ due to spatial variation.
\end{itemize}

A necessary and sufficient condition for straight-line trajectories is
\[
\frac{Dv}{Dt} \equiv 0.
\]

\subsection*{A.2 Relation to One-Step Integration Error}

Consider a Taylor expansion of the trajectory around $t=0$:
\[
x(1)
=
x_0
+
v(x_0,0)
+
\frac{1}{2} \frac{Dv}{Dt}(x_0,0)
+
\mathcal{O}(\|Dv/Dt\|^2).
\]

The one-step Euler approximation is
\[
\hat{x}_1 = x_0 + v(x_0,0).
\]

Thus, the leading-order integration error satisfies
\begin{equation}
\label{eq:euler_error}
\|x(1) - \hat{x}_1\|
\;\approx\;
\frac{1}{2}
\left\|
\frac{Dv}{Dt}(x_0,0)
\right\|.
\end{equation}

\textbf{Implication.}
Accurate few-step or one-step generation requires suppressing the material acceleration of the learned velocity field, particularly at early times.

---

\subsection*{A.3 Self-Guided Finite-Difference Approximation}

Direct evaluation of $\frac{Dv}{Dt}$ requires Jacobian–vector products, which are computationally expensive for large neural networks.

Iso-FM replaces this with a \emph{self-guided finite-difference estimator}.
Given a small $\varepsilon > 0$, define a forward Euler step using the model’s own velocity:
\[
x_{t+\varepsilon}
=
x_t
+
\varepsilon \, v_\theta(x_t,t).
\]

A first-order Taylor expansion of the velocity yields
\[
v_\theta(x_{t+\varepsilon}, t+\varepsilon)
=
v_\theta(x_t,t)
+
\varepsilon
\left(
\partial_t v_\theta
+
(v_\theta \cdot \nabla_x) v_\theta
\right)
+
\mathcal{O}(\varepsilon^2).
\]

Rearranging,
\begin{equation}
\label{eq:finite_diff}
\frac{
v_\theta(x_{t+\varepsilon}, t+\varepsilon)
-
v_\theta(x_t,t)
}{\varepsilon}
\;\approx\;
\frac{Dv_\theta}{Dt}(x_t,t).
\end{equation}

Therefore, minimizing
\[
\|v_\theta(x_t,t) - v_\theta(x_{t+\varepsilon}, t+\varepsilon)\|^2
\]
acts as a stochastic, Jacobian-free surrogate for minimizing the squared material acceleration.

This is the core principle behind \textbf{self-guided Isokinetic Flow Matching}.

\subsection*{A.4 Effect on Flow Geometry}

If Iso-FM successfully enforces
\[
\frac{Dv_\theta}{Dt} \approx 0
\quad \text{along trajectories},
\]
then velocity is approximately constant along each path:
\[
v_\theta(x(t),t) \approx v_\theta(x_0,0).
\]

Integrating the ODE yields
\[
x(t) \approx x_0 + t \, v_\theta(x_0,0),
\]
i.e., the flow map becomes approximately linear in time.

\textbf{Key insight.}
Iso-FM does not learn flow maps explicitly; instead, it reshapes the Eulerian field so that the induced Lagrangian maps are simple.

\subsection*{A.5 The Fundamental Limit of Flow Straightening}
\label{app:fundamental_limit}

While Iso-FM successfully minimizes the material acceleration $\frac{Dv}{Dt}$ to straighten the marginal velocity field, achieving a perfectly linear global Lagrangian flow map (where $\frac{Dv}{Dt} \equiv 0$ everywhere) is mathematically impossible when matching a simple prior to a complex, multimodal data distribution. We demonstrate this analytically by connecting the marginal acceleration to the conditional velocity variance.

Let $z$ be a generalized conditioning variable (e.g., a specific target $z=x_1$, or a coupled source-target pair $z=(x_0, x_1)$). Let $p(x,t|z)$ and $u(x,t|z)$ be the conditional probability density and conditional velocity field, respectively. By construction in Optimal Transport Flow Matching and Rectified Flow, the conditional paths are straight, meaning their material acceleration is strictly zero:
\begin{equation}
\label{eq:cond_acceleration_zero}
\frac{Du}{Dt} = \partial_t u + (u \cdot \nabla_x) u = 0.
\end{equation}
The marginal density $p(x,t)$ and marginal velocity $v(x,t)$ are defined via marginalization over the prior distribution $q(z)$:
\begin{align}
p(x,t) &= \int p(x,t|z) q(z) dz, \\
v(x,t) &= \frac{1}{p(x,t)} \int u(x,t|z) p(x,t|z) q(z) dz = \mathbb{E}_{z|x_t=x}[u].
\end{align}
Both the conditional and marginal fields satisfy their respective continuity equations: $\partial_t p(\cdot|z) + \nabla_x \cdot (p(\cdot|z) u) = 0$ and $\partial_t p + \nabla_x \cdot (p v) = 0$.

We seek to evaluate the material acceleration of the marginal field, $\frac{Dv}{Dt} = \partial_t v + (v \cdot \nabla_x) v$. We begin by taking the partial time derivative of the marginal momentum $p v$:
\begin{equation}
\partial_t (p v) = \int \partial_t \big( p(\cdot|z) u \big) q(z) dz = \int \big( u \partial_t p(\cdot|z) + p(\cdot|z) \partial_t u \big) q(z) dz.
\end{equation}
Substituting the conditional continuity equation $\partial_t p(\cdot|z) = -\nabla_x \cdot (p(\cdot|z) u)$, we get:
\begin{equation}
\partial_t (p v) = \int \Big( -u \nabla_x \cdot (p(\cdot|z) u) + p(\cdot|z) \partial_t u \Big) q(z) dz.
\end{equation}
Using the vector calculus identity $\nabla_x \cdot (p u u^T) = u \nabla_x \cdot (p u) + p (u \cdot \nabla_x) u$, we can rewrite the integrand:
\begin{equation}
\partial_t (p v) = \int \Big( -\nabla_x \cdot (p(\cdot|z) u u^T) + p(\cdot|z) \big[ \partial_t u + (u \cdot \nabla_x) u \big] \Big) q(z) dz.
\end{equation}
By Equation~\ref{eq:cond_acceleration_zero}, the bracketed term is exactly zero. Pushing the integral inside the divergence yields:
\begin{equation}
\partial_t (p v) = -\nabla_x \cdot \left( p(x,t) \mathbb{E}_{z|x_t=x}[u u^T] \right).
\end{equation}
We now apply the Reynolds decomposition from fluid mechanics. The expectation of the uncentered second moment of the conditional velocities can be decomposed into the outer product of the marginal mean velocity plus the conditional covariance matrix $\Sigma(x,t)$:
\begin{equation}
\mathbb{E}_{z|x_t=x}[u u^T] = v v^T + \Sigma(x,t), \quad \text{where} \quad \Sigma(x,t) = \mathbb{E}_{z|x_t=x}[(u - v)(u - v)^T].
\end{equation}
Substituting this decomposition back into the momentum equation yields:
\begin{equation}
\partial_t (p v) = -\nabla_x \cdot (p v v^T) - \nabla_x \cdot (p \Sigma).
\end{equation}
Expanding the left side using the product rule ($\partial_t(pv) = v\partial_t p + p\partial_t v$) and the first term on the right side ($\nabla_x \cdot (p v v^T) = v \nabla_x \cdot (p v) + p (v \cdot \nabla_x) v$), we obtain:
\begin{equation}
v \big[ \partial_t p + \nabla_x \cdot (p v) \big] + p \big[ \partial_t v + (v \cdot \nabla_x) v \big] = -\nabla_x \cdot (p \Sigma).
\end{equation}
The first bracketed term is identically zero due to the marginal continuity equation. The second bracketed term is exactly the material derivative of the marginal velocity. Dividing by $p(x,t)$, we arrive at the fundamental limit equation:
\begin{equation}
\label{eq:marginal_acceleration_variance}
\frac{Dv}{Dt}(x,t) = -\frac{1}{p(x,t)} \nabla_x \cdot \big( p(x,t) \Sigma(x,t) \big).
\end{equation}

\textbf{Physical Implication.} 
Equation~\ref{eq:marginal_acceleration_variance} proves that the marginal acceleration $\frac{Dv}{Dt}$ is strictly governed by the spatial divergence of the conditional velocity variance $\Sigma(x,t)$ (the ``kinetic stress tensor"). At early times ($t \approx 0$), the prior $p_0$ is highly uninformative, meaning a single spatial coordinate $x$ has a non-zero probability of routing to widely separated data modes. Thus, the posterior variance $\Sigma(x,0)$ is strictly positive. 

Because the probability density $p(x,t)$ must decay to zero at spatial infinity ($\|x\| \to \infty$), the non-negative tensor field $p(x,t)\Sigma(x,t)$ must also vanish at infinity. A smooth field that decays to zero at boundaries cannot have a globally zero divergence unless the field itself is identically zero. Therefore, $\nabla_x \cdot (p\Sigma)$ cannot be globally zero, meaning the material acceleration $\frac{Dv}{Dt}$ \emph{must} be non-zero. 

Physically, a particle drawn from the prior must accelerate to ``commit" to a specific data mode as $t \to 1$ and the posterior collapses. Therefore, while Iso-FM successfully suppresses arbitrary convective acceleration, a perfectly linear global transport map is theoretically unachievable unless the data-to-noise coupling is entirely deterministic and one-to-one ($\Sigma \equiv 0$), which violates the independent or minibatch-OT coupling formulations of standard generative flows.

\section*{Appendix B: Relation to Flow Maps and Solution Operators}

\subsection*{B.1 Eulerian vs.\ Lagrangian Objectives}

Let $\Phi_t$ denote the solution operator (flow map) induced by $v$:
\[
\Phi_t(x_0) = x(t).
\]

\begin{itemize}
\item \textbf{Eulerian learning} (Flow Matching): learn $v(x,t)$ locally.
\item \textbf{Lagrangian learning} (Flow-map / consistency methods): learn $\Phi_t$ directly.
\end{itemize}

Learning $\Phi_t$ collapses the ODE into a direct mapping but removes intermediate-time semantics and requires additional supervision or distillation.

Iso-FM remains Eulerian but modifies the learned field so that $\Phi_t$ becomes trivial to approximate numerically.

---

\subsection*{B.2 Learning Velocity and Flow Maps Jointly (Preliminary)}

Motivated by the equivalence
\[
\partial_t \Phi_t(x) = v(\Phi_t(x), t),
\]
one may consider learning both:
\begin{itemize}
\item a velocity field $v_\theta(x,t)$,
\item a flow map $\Phi_\psi(x,t)$,
\end{itemize}
with a consistency constraint
\[
\Phi_\psi(x_t,t)
\;\approx\;
\Phi_\psi(x_t + \varepsilon v_\theta(x_t,t), t+\varepsilon).
\]

In this view:
\begin{itemize}
\item Iso-FM regularizes the \emph{Eulerian dynamics},
\item the flow-map head learns the corresponding \emph{Lagrangian operator}.
\end{itemize}

Preliminary toy experiments suggest that Iso-FM significantly stabilizes such joint training by reducing stiffness of the velocity field. However, we emphasize that this direction remains exploratory and is not required for few-step gains reported in the main paper.

\section*{Appendix C: Limitations and Open Problems}

While Isokinetic Flow Matching (Iso-FM) provides a principled and lightweight mechanism for reducing trajectory curvature and enabling efficient few-step sampling, several limitations and open questions remain.

\subsection*{C.1 Limits of One-Step Generation}

Iso-FM substantially reduces material acceleration of the learned velocity field, which in turn lowers discretization error for coarse numerical solvers. However, this alone does not guarantee high-fidelity \emph{one-step} generation in complex, multimodal data distributions.

In particular, when the conditional posterior $p(x_1 \mid x_0)$ is strongly multimodal at early times, the optimal marginal velocity remains an average over competing directions. While Iso-FM suppresses rapid changes in this average along trajectories, it does not resolve the fundamental ambiguity at $t=0$. Empirically, this manifests as a performance gap between $1$-step and $2$-step sampling that persists even under strong regularization.

This limitation is intrinsic to instantaneous velocity regression and suggests that true one-step generation may require either:
(i) explicit flow-map learning,
(ii) additional symmetry-breaking mechanisms,
or (iii) auxiliary conditioning information.

\subsection*{C.2 No Guarantee of Global Non-Intersection}

Iso-FM enforces low acceleration \emph{locally along trajectories}, but does not impose a global constraint preventing different trajectories from intersecting in state space.

Although intersections become exponentially unlikely in high-dimensional spaces, particularly for image data, a formal guarantee of global non-intersection remains open. Understanding whether additional weak regularization terms could promote global topological consistency without sacrificing efficiency is an important direction for future work.

\subsection*{C.3 Hyperparameter Sensitivity}

The effectiveness of Iso-FM depends on several hyperparameters, including the isokinetic weight $\lambda$, the time-weighting exponent $\alpha$, and the distribution used to sample the lookahead step $\varepsilon$.

While we find that log-normal or Beta-distributed $\varepsilon$ schedules are robust across tasks, a principled method for selecting these parameters remains an open problem. Adaptive schemes based on local curvature, posterior entropy, or velocity variance may offer a more automatic alternative.

\subsection*{C.4 Relation to Flow-Map Learning}

Iso-FM modifies the Eulerian dynamics so that the induced Lagrangian flow maps become approximately linear and easy to integrate numerically. However, Iso-FM does not explicitly learn these maps.

Recent work on flow-map learning and consistency models demonstrates that directly learning solution operators can achieve true one-step generation. A precise theoretical characterization of when Iso-FM suffices to recover these operators implicitly—and when explicit flow-map learning is necessary—remains unresolved.

Our preliminary toy experiments suggest that combining Iso-FM with lightweight flow-map heads may be promising, but we leave a systematic investigation of this hybrid regime to future work.

\subsection*{C.5 Theoretical Generalization Bounds}

While we provide local Taylor-based arguments linking Iso-FM to reduced Euler integration error, global generalization guarantees for long-time transport remain open.

In particular, bounding the accumulated error of few-step solvers as a function of the expected material acceleration along learned trajectories is an important theoretical question that may connect Iso-FM to recent advances in neural ODE stability analysis.

\subsection*{C.6 Scalability to Very Large Models}

Iso-FM introduces minimal memory overhead and avoids second-order derivatives, making it compatible with large architectures such as DiT. However, the interaction between isokinetic regularization and optimization dynamics at scale—especially under mixed-precision training and aggressive optimizer settings—requires further study.

Understanding how Iso-FM interacts with architectural inductive biases, attention mechanisms, and normalization layers is another promising direction.

\subsection*{C.8 Broader Perspective}

Iso-FM represents a step toward \emph{dynamically informed generative modeling}, where local physical constraints on motion shape global generative behavior. We believe that future work at the intersection of generative modeling, numerical analysis, and dynamical systems will continue to yield both theoretical insights and practical speedups.

\subsection*{Takeaway}

Iso-FM provides a principled mechanism for straightening generative flows at the dynamical level.
Once acceleration is suppressed, learning an explicit solution operator becomes optional rather than necessary.

In this sense, Iso-FM bridges Eulerian and Lagrangian generative modeling by enforcing conditions under which they coincide.

\section*{Appendix D: Additional Generated Samples}

\begin{figure}[htbp]
\centering
\includegraphics[width=0.48\linewidth]{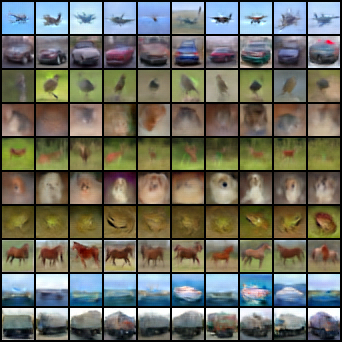}
\hfill
\includegraphics[width=0.48\linewidth]{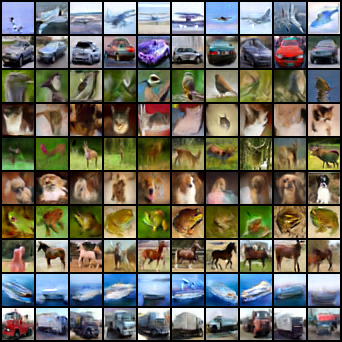}
\caption{Conditional samples: FM (left, epoch 1250) vs Iso-FM (right, epoch 1250).}
\label{fig:appendix_conditional_samples}
\end{figure}

\begin{figure}[htbp]
\centering
\includegraphics[width=0.48\linewidth]{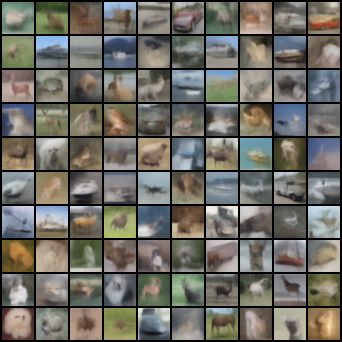}
\hfill
\includegraphics[width=0.48\linewidth]{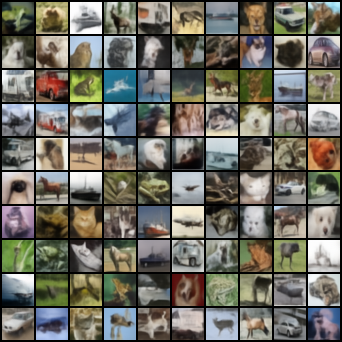}
\caption{Unconditional OT samples: OT+FM (left, epoch 500) vs OT+Iso-FM (right, epoch 1250).}
\label{fig:appendix_unconditional_ot_samples}
\end{figure}

\end{document}